\documentclass[submission,copyright,creativecommons]{eptcs}
\providecommand{\event}{SOS 2007} 
\usepackage{breakurl}             
\usepackage{underscore}           

\usepackage{times}
\usepackage{soul}
\usepackage{url}
\usepackage[utf8]{inputenc}
\usepackage[small]{caption}
\usepackage{graphicx}
\usepackage{amsmath}
\usepackage{amssymb}
\usepackage{amsfonts}
\usepackage{amsthm}
\usepackage{booktabs}
\usepackage[ruled,lined]{algorithm2e}

\usepackage[inline]{enumitem}
\usepackage{tikz}
\usepackage{xspace}
\usepackage{listings}
\usepackage{paralist}

\usepackage{multirow}
\usepackage{lscape}
\usepackage[figuresright]{rotating}
\usepackage{afterpage}

\newlist{inlineitemize}{enumerate*}{1}
\setlist[inlineitemize]{label=(\roman*)}

\graphicspath{ {./img/} }

\newcommand{\finishes}{\ensuremath{\langle E \rangle}\xspace}
\newcommand{\during}{\ensuremath{\langle D \rangle}\xspace}
\newcommand{\starts}{\ensuremath{\langle B \rangle}\xspace}
\newcommand{\overlaps}{\ensuremath{\langle O \rangle}\xspace}
\newcommand{\meets}{\ensuremath{\langle A \rangle}\xspace}

\newcommand{\later}{\ensuremath{\langle L \rangle}\xspace}

\newcommand{\mmodels}{\Vdash}

\title{Decision Tree Learning with Spatial Modal Logics}
\author{%
Giovanni Pagliarini
\institute{Dept. of Mathematics and Computer Science\\University of Ferrara, Italy}
\institute{Dept. of Mathematical, Physical and Computer Sciences\\University of Parma, Italy}
\email{pglgnn@unife.it}
\and
Guido Sciavicco
\institute{Dept. of Mathematics and Computer Science\\University of Ferrara, Italy}
\email{scvgdu@unife.it}
}
\def\titlerunning{Decision Tree Learning with Spatial Modal Logics}
\def\authorrunning{G. Pagliarini, \& G. Sciavicco}
\begin{document}
\maketitle

\begin{abstract}
Symbolic learning represents the most straightforward approach to interpretable modeling,
but its applications have been hampered by a single structural design choice: the adoption of propositional logic as the underlying language.
Recently, more-than-propositional symbolic learning methods have started to appear, in particular for time-dependent data. These methods exploit the expressive power of
modal temporal logics in powerful learning algorithms, such as temporal decision trees, whose classification capabilities are comparable with the best non-symbolic ones, while
producing models with explicit knowledge representation. With the intent of following the same approach in the case of spatial data, in this paper we: \begin{inlineitemize} \item present a theory of spatial decision tree learning; \item describe a prototypical implementation of a spatial decision tree learning algorithm based, and strictly extending, the classical C4.5 algorithm, and
\item perform a series of experiments in which we compare the predicting power of spatial decision trees with that of classical propositional decision trees in several versions, for a multi-class image classification problem, on publicly available datasets. \end{inlineitemize} Our results are encouraging, showing clear improvements in the performances from the propositional to the spatial models, which in turn show higher levels of interpretability.
\end{abstract}


\section{Introduction}
{\em Connectionist} learning has been the driving force of machine learning for at least a decade, allowing advances in many difficult artificial intelligence tasks. Recently, however, we have seen a rising need for models that are interpretable and explainable, and it has been pointed out that, despite their potential and versatility, neural models often enclose complex algebraic functions that are difficult to interpret; such a need is even reflected in recent political initiatives (see, e.g., the 2016 General Data Protection Regulation - GDPR - of the European Union). A well-known approach to interpretable modeling is based on replacing the algebraic language that underlies a model with a logical language, and this foundational choice leads to the field of {\em symbolic learning}. As opposed to functional learning, where the learned model is an algebraic function, symbolic models are structures of logical formulas that are directly translatable to natural language. While logical languages make up a whole research field, in which languages are studied and categorized according to their expressive and computational properties, symbolic learning has always been limited to the use of propositional and first-order languages. Symbolic learning at the propositional level is very well-known, and includes a plethora of widely used methods ranging from {\em decision trees} to {\em rule-based models}; their clearest limit is the fact that complex relationships in data cannot be directly expressed. Symbolic learning at the first-order level is known as {\em inductive logic programming} (or ILP), and it is based on the assumption that entities and relationships in data are explicit. The limit of inductive logic programming is the fact that the extracted models are often very complex, and generally expressed in terms of logic programs, which is not suitable in many practical cases.


\smallskip

An alternative to both propositional and first-order symbolic learning is {\em modal symbolic learning}, which is emerging in the past few years. The idea is to re-design and generalize classical symbolic algorithms to exploit the expressive power of {\em modal logics}~\cite{ModalLogics}, which are more expressive than propositional ones but less than first-order ones. Modal logics are tailored to describe dimensional data in which the dimension(s) are implicit. Typical examples of modal logics are {\em temporal} and {\em spatial} logics. {\em Temporal symbolic learning} has been first presented in this form in~\cite{Brunello2019,Extraction,KnowledgeExtractionTemporalSymbolic}, 
where the authors study the use of a temporal modal logic (in particular, {\em Halpern and Shoham's logic of time intervals}, or $\mathcal{HS}$~\cite{HS}) in order to learn temporal decision trees for multivariate time-series classification. Among the several examples of multivariate time-series classification models that can be extracted with these techniques in the form of trees or sets of rules, consider the illuminating example of the medical context, in which patients under observation for a period of time may be classified by patterns of the form {\em if there exists an interval in time in which the body temperature is higher than $39$, overlapped by an interval where heart rate is higher than $140$ then the patient has a certain condition.} The interval-based approach turns out to be very natural, allowing one to make decisions that are essentially not point-based (in our example, evaluating the temperature of a patient at a single moment is not really significative). Although this idea is still in its infancy, the initial results seem to indicate that the learned models are competitive, in statistical terms, with those that belong to more classical approaches (both symbolic and functional), while showing a high level of interpretability.

\smallskip
In this paper, we apply the same approach to the spatial domain by extending a classical decision tree learning algorithm, namely C4.5~\cite{Quinlan93}, to allow it to pick up spatial patterns and describe them in terms of spatial relations between entities in a geometric space. While interval logic is the most natural choice to describe extended patterns in the single-dimensional (i.e., temporal) case, a wide range of possibilities arises with more than one dimensions: in fact, there exist several spatial logics characterized by their underlying ontology, represented by the choice of what constitutes spatial {\em entities} (e.g., triangles, rectangles, generic areas, and so on) and {\em relations} (e.g., {\em containing}, {\em next to}, {\em to the right of}, and so on). However, it turns out that generalizing interval temporal logic to many dimensions captures many of these choices, and it presents itself as a suitable compromise between a simple enough syntax and a high enough expressive power. Following the same interval-based approach as in the temporal case, in two dimensions intervals are replaced by rectangles, and our models can learn to classify an image according to whether it displays a pattern such as, for example, {\em there exists a region in the image with a high level of red, containing another region with a low level of green}. Moreover, such an approach can be further generalized to many dimensions, which may be useful for learning from multi-dimensional data (e.g., hyperspectral video streams). After defining a generalized version of $\mathcal{HS}$, we test our implementation in the bi-dimensional case on a {\em land cover classification} problem, using public benchmark datasets. Since in a typical setting models for this problem are required to be interpretable, land cover classification has been initially approached using (propositional) decision tree learning~\cite{Corn,InferringMethodLand}; nevertheless, in recent times functional methods seem to have taken over~\cite{DBLP:journals/corr/CaoZXMXP17,DBLP:journals/tgrs/SantaraMHSGPM17}. Since functional methods are specifically designed to obtain the best statistical performances, even at the expense of the interpretability of the model, they are not immediately comparable with symbolic ones, and such a comparison would probably not be very significant. In this paper, we show how the spatial symbolic approach improves both the performances and the expressive power of the propositional one, and how a logical theory of the underlying phenomenon can be extracted,  discussed, interpreted, and translated into natural language,
which would not be possible with a functional method. Although they are not immediately comparable to our solution, which is part of a long-term project that aims at generalizing symbolic learning with modal logics in a comprehensive way, there have been attempts at logic-based spatial learning. Among them we mention~\cite{10.1007/1156412620}, in which the authors extract a regression function that depend on the spatial relationships among objects in an image, and it is more database-oriented rather than a pure learning method, and~\cite{cohn}, in which a customized ILP-like approach is used to extract knowledge from video instances (in some sense, this approach is vaguely comparable with ours, except it requires first-order description of the data before the learning can take place). Finally, in~\cite{BLOCKEEL1998285} the authors propose a methodology to extract first-order decision trees; such an approach is comparable with ours (as we extract propositional modal decision trees), but, once again, in their case relations are supposed to be explicit, so the two approaches cannot be immediately applied to the same situations.

\smallskip

This paper is organized as follows. In Section~\ref{sec:logic} we briefly review the most important ontologies of space and present a generalization of the interval temporal logic $\mathcal{HS}$ for multi-dimensional reasoning. In Section~\ref{sec:dt-learning} we present a formalization of classical decision tree learning, and its generalization to the case of modal logics. Then, in Section~\ref{sec:exp-results} we present the {\em land cover classification} problem, and we evaluate the performances of spatial decision trees on this problem, using public datasets, before concluding.
%



\begin{figure*}[t]
\centering
\begin{tikzpicture}[scale=.8]
 \tikzstyle{every node}=[font=\footnotesize]

\draw (0,0)node(op){\bf $\mathcal{HS}$};

%

\draw (op) ++(0,-1.5)node(meets){$\meets$};
\draw (meets)++(0,-.75)node(later){$\later$};
\draw (meets)++(0,-1.5)node(starts){$\starts$};
\draw (meets)++(0,-2.25)node(finishes){$\finishes$};
\draw (meets)++(0,-3)node(during){$\during$};
\draw (meets)++(0,-3.75)node(overlaps){$\overlaps$};


\draw (meets)++(1.7,0) node(sep1){} ++(0,.5) -- ++(0,-4.7);


\draw (sep1)++(3,1.5)node(rel){\bf Allen's relations};

\draw (sep1)++(0.5,0)node[right](Ra){$[x,y] R_A [x',y'] \Leftrightarrow y=x'$};
\draw (sep1)++(0.5,-.75)node[right](Rl){$[x,y] R_L [x',y'] \Leftrightarrow y < x'$};
\draw (sep1)++(0.5,-1.5)node[right](Rs){$[x,y] R_B [x',y'] \Leftrightarrow x=x', y' < y$};
\draw (sep1)++(0.5,-2.25)node[right](Rf){$[x,y] R_E [x',y'] \Leftrightarrow y=y', x < x'$};
\draw (sep1)++(0.5,-3)node[right](Rd){$[x,y] R_D [x',y'] \Leftrightarrow x < x', y' < y$};
\draw (sep1)++(0.5,-3.75)node[right](Ro){$[x,y] R_O [x',y'] \Leftrightarrow x < x' < y < y'$};


\draw (sep1)++(8.5,0)node(sep2){} ++(0,0.5) -- ++(0,-4.7);

\draw (sep2)++(2.8,1.5)node(graphic){\bf Graphical representation};

\draw[red,|-|] (sep2) ++(.7,.75)node[above](a){$x$} -- ++(2,0)node[above](b){$y$};
\draw[dashed,red,help lines,thick] (a) -- ++(0,-5.25);
\draw[dashed,red,help lines,thick] (b) -- ++(0,-5.25);

\draw[|-|] (b) ++(0,-1) ++(0,0)node[above](Ac){$x'$}
-- ++(1,0)node[above](Ad){$y'$};
\draw[|-|] (b) ++(0,-1) ++(.5,-.75)node[above](Lc){$x'$}
-- ++(1,0)node[above](Ld){$y'$};
\draw[|-|] (a) ++(0,-1) ++(0,-1.5)node[above](Bc){$x'$}
-- ++(.5,0)node[above](Bd){$y'$};
\draw[|-|] (b) ++(0,-1) ++(-.5,-2.25)node[above](Ec){$x'$}
-- ++(.5,0)node[above](Ed){$y'$};
\draw[|-|] (a) ++(0,-1) ++(.5,-3)node[above](Dc){$x'$}
-- ++(1,0)node[above](Dd){$y'$};
\draw[|-|] (a) ++(0,-1) ++(1,-3.75)node[above](Oc){$x'$}
-- ++(2,0)node[above](Od){$y'$};
\end{tikzpicture}


\caption{Allen's interval relations and Halpern and Shoham's ($\mathcal {HS}$) modalities; $\mathcal{HS}$ is the single-dimensional case of modal logic of hyperrectangles.}

\label{fig:relations}
\end{figure*}

\section{Modal Logics of Space} \label{sec:logic}

\noindent{\bf Ontologies of space.} Because {\em qualitative spatial reasoning} (QSR) has important real-world applications (e.g., in geographic information systems), there is active research in deriving formal systems for reasoning about space.
Throughout the last 30 years, many formalizations have been proposed for different purposes~\cite{ConstraintCalculiQSR2007}, mostly in the form of {\em binary spatial calculi}, but sometimes also in logical terms.
As mentioned, two major structural choices for deriving a spatial logic concern the definition of {\em spatial entities}, namely, regions, and the set of relevant {\em binary relations} between such entities. These choices are to be tailored to the domain at hand. Depending on the application, objects are classically chosen to be represented as points or regions of a specific kind (e.g., convex regions).
Points are more elementary, and they cannot express the extension of objects, thus they are not suitable in contexts where objects have diverse shapes, or when mereological aspects of space are relevant (e.g., grade of overlap between regions).
Regions, on the other hand, can express extension, but are more complex to deal with.
When regions are considered, they are usually assumed to be {\em connected},
which is in agreement with an implicit locality hypothesis on the objects of the reasoning. Although limited in terms of expressive power, convex regions are typically preferred as extended entities, or they are used to approximate complex shapes (e.g., by means of {\em convex hulls}~\cite{DBLP:conf/cosit/Cohn95}). Bidimensional convex regions providing a good trade-off between complexity of reasoning and expressive power include {\em triangles} and {\em convex quadrilaterals}. Regions can be further constrained in order to achieve higher specificity or lower complexity; for example, forcing all entities to be of the same size can be beneficial for specific domains (e.g., {\em optical character recognition}). Another sensible constraint is {\em orthogonality}: in the case of polygons, forcing the edges to be parallel to a given reference set of axes limits the set of possible relations, and sometimes lowers computational costs; in particular, rectangles with edges parallel to the axes have been deeply used in QSR due to their computational benefits. As for relation sets, different aspects can be considered (e.g., distance, size, shape), and two of the most studied relational aspects are {\em directionality} and {\em topology}. Directional relations account for the relative spatial arrangement of the two objects (e.g., {\em next to}, {\em to the right of}); they are not invariant under rotation/reflection, and are not as easily definable for generic types of regions. On the algebraic side, many calculi have been proposed, both with punctual entities~\cite{cardinal-relations} and regions of various types~\cite{DBLP:conf/kr/BalbianiCC98,RectCDC,CDC,ConeShapedExtended}; directional algebras could, in principle, give rise to their modal logic counterparts, but only a few attempts have been made in this sense~\cite{DBLP:conf/time/BresolinSMMS10,CompassLogic,ConeLogic,SpPNL}. While in this context an absolute frame of reference is generally adopted, there are examples of spatial logics and calculi that account for the {\em orientation} of objects, and are able to describe relations such as {\em in front of}, {\em behind} or {\em facing each other}~\cite{using-freksa,SOSL}. On the other hand, topological relations generally address the different modalities of intersection between regions and their boundaries, which makes them invariant under rotation and reflection; being purely qualitative and easily definable for generic regions, topology is the most common choice, and is often preferred to directional approaches. The most popular characterization of topological binary relations is the set of Egenhofer-Franzosa relations, often referred to as RCC8~\cite{49-IM,RCC8}. In its extended form, the set includes the $8$ relations: {\em disconnected}, {\em externally connected}, {\em partially overlapping}, {\em tangential proper part}, {\em non-tangential proper part}, {\em tangential proper part inverse}, {\em non-tangential proper part inverse}, and {\em equality}. RCC8 can be made coarser by using unions of relations; the natural resulting, smaller, set of relations is known as RCC5. In~\cite{LRCC8} the modal logics that are entailed by these sets of relations, namely $\mathcal L_{RCC8}$ and $\mathcal L_{RCC5}$, are studied. Compared to topological relations, {\em jointly exhaustive } region-based directional algebras (namely, in which each pair of entities are in at least a relation) generally present a high number of relations.

%

\smallskip

\noindent{\bf A spatial logic for learning}. Learning is an inductive process. As such, the logical language on which a learning model is designed is interesting for properties other than deduction ones. Inspired by both Halpern and Shoham's interval temporal logic~\cite{HS} and Rectangle Algebra~\cite{DBLP:conf/kr/BalbianiCC98}, here we introduce a very general spatial logic based on hyperrectangles on a finite, multi-dimensional space. Let:
$$\mathbb D=\langle D,<\rangle = [1, \ldots, N]$$

\noindent be a finite, linearly ordered set, and let $\mathbb D^k=\mathbb D\times\ldots\times\mathbb D$ be a $k$-dimensional, finite and discrete geometrical space. Elements of $\mathbb D^k$ are called points, and are denoted as $\pi=(x_1,\ldots,x_k)$. In analogy with interval temporal logic, an {\em hyperrectangle} in $\mathbb D^k$ is an object of the type:
$$[(x_1,y_1),(x_2,y_2),\ldots,(x_k,y_k)],$$

\noindent where for each $i$, $x_i<y_i$ (observe that, using this notation, in the one-dimensional case hyperrectangles are just intervals in their familiar notation, that is, $[x_1,y_1]$, by simply omitting the inner brackets). Hyperrectangles are essentially the extension of intervals in a higher dimensional space; as such, the notion of $\pi$ {\em belonging} to a hyperrectangle ($\pi\in[(x_1,y_1),(x_2,y_2),\ldots,(x_k,y_k)]$) can be defined using the ordering relation of $\mathbb D$, and, in fact, different such notions can be defined depending on the particular application. Now, let $\mathbb K(\mathbb D^k)$ be the set of all hyperrectangles that can be formed on $\mathbb D^k$; in the following, we use $r,s\ldots$ to denote hyperrectangles.
On a linear order, there are 13 distinct relations between any two intervals; these are commonly referred to as \textit{Allen's relations}~\cite{Allen}, and the set is given by the six relations {\em After}, {\em Later}, {\em Begins}, {\em Ends}, {\em During}, {\em Overlaps}, their inverses, and the {\em equality} relation. Note that all except for the equality are logically interesting, thus in a single dimension, we represent relations by a symbol $R_X$, where $X\in\{A,L,B,E,D,O,\overline{A},\overline{L},\overline{B},\overline{E},\overline{D},\overline{O}\}$; their informal semantics is depicted in Fig.~\ref{fig:relations}, in which we used Halpern and Shoham's notation.
In the multi-dimensional generalization, we represent any relation as a tuple of $k$ single-dimension relations, that is, $(R_{X_1},\ldots,R_{X_k})$, with $X_i\in\{A,L,B,E,D,O,\overline{A},\overline{L},\overline{B},\overline{E},\overline{D},\overline{O},=\}$, with the additional constraint that they cannot be all the equality (since, again, equality cannot be lifted to the level of logical modality).
Ultimately, this leads to a number of $13^k-1$ distinct relations between any two hyperrectangles in a $k$-dimensional space.
Lifting such relations at the modal level entails introducing unary modal operators of the type:
$$\langle X_1,\ldots,X_k\rangle,$$

\noindent where, again, each $X_i\in\{A,L,B,E,D,O,\overline{A},\overline{L},\overline{B},\overline{E},\overline{D},\overline{O},=\}$, and where the operator $\langle =,\ldots,=\rangle$ is excluded. Formulas of the {\em modal logic of hyperrectangles}, denoted $\mathcal {HS}^k$, are therefore obtained by the following grammar:
$$ \varphi ::= p \mid \neg \varphi \mid \varphi \lor \varphi \mid \langle X_1,\ldots,X_k \rangle \varphi,$$

\noindent where $p$ is a propositional letter from a given set $\mathcal{P}$. Such formulas are easily interpreted in a $k$-{\em dimensional spatial model} $\mathcal M=\langle \mathbb K(\mathbb D^k),V\rangle$, where $V$ is an evaluation function defined as:
$$V:\mathbb K(\mathbb D^k)\mapsto 2^{\mathcal P}$$

\noindent which assigns to each hyperrectangle the set of all and only propositional letters that are true on it, by means of the following truth relation:
$$
\begin{array}{lll}
\mathcal M, r \mmodels p                              &   \text{if}  &       p \in V(r) \\
\mathcal M, r \mmodels \neg \varphi                   &   \text{if}  &       \mathcal M, r \not\mmodels \varphi \\
\mathcal M, r \mmodels \varphi \lor \psi              &   \text{if}  &       \mathcal M, r \mmodels \varphi \text{ or } \mathcal M, r \mmodels \psi \\
\mathcal M, r \mmodels \langle X_1,\ldots,X_k \rangle \varphi      &   \text{if}  &  \exists s \text{ s.t. } r(R_{X_1},\ldots,R_{X_k})s,\ \mathcal M, s \mmodels \varphi, \\
\end{array}
$$

\noindent where, $r \in \mathbb K(\mathbb D^k)$.
The universal version $[X_1,\ldots,X_k]$ of an existential modality is defined  in the standard way. When $k=1$ the resulting logic is precisely Halpern and Shoham's $\mathcal {HS}$, and when $k=2$ we obtain the natural unary modal logic that emerges from Rectangle Algebra. Moreover, for any $k$, one can easily define derived relations (and thus, modal operators) through union (that is, disjunction); for example, for $k=2$, the topological relation {\em disconnected} (usually denoted $DC$) can be defined as the following union:
$$\bigcup_{X\in\{A,L,B,E,D,O,=\}} (\overline L,X) \cup (L,X) \cup (X,\overline L) \cup (X,L).$$

\begin{figure}[t]
		\centering
		\includegraphics[width=.85\linewidth]{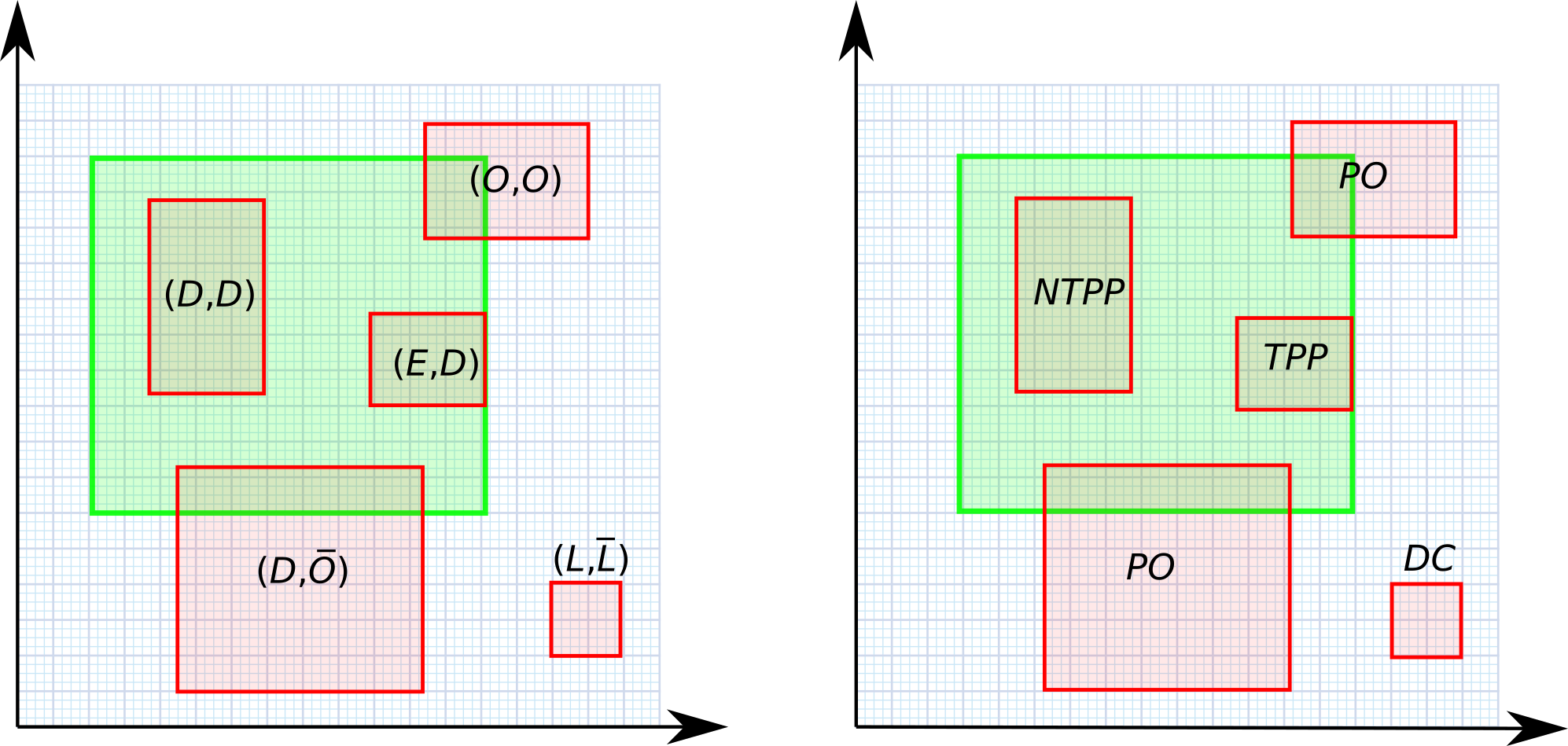}
		\caption{Examples of $\mathcal {HS}$$^2$ relations (left) and of $\mathcal {HS}$$^2_{RCC8}$ ones (right); in both cases, the green rectangle is the reference.}
		\label{fig:dir-top}
\end{figure}

\noindent In a similar way, each of the RCC8 and RCC5 relations can be expressed as disjunctions of $\mathcal {HS}^k$ operators; in fact, restricting the language to RCC8 or RCC5 modalities corresponds to obtaining Lutz and Wolter's $\mathcal{L}_{RCC8}$ and $\mathcal{L}_{RCC5}$ interpreted on rectangles. In these two last cases, we use the names $\mathcal {HS}^2_{RCC8}$ and $\mathcal {HS}^2_{RCC5}$. While encompassing as much as 168 different modal operators, $\mathcal{HS}^2$ is very intuitive; a pictorial example of the relations that can be expressed in $\mathcal{HS}^2$ is shown in Fig.~\ref{fig:dir-top} (left). $\mathcal {HS}^2_{RCC8}$ and $\mathcal {HS}^2_{RCC5}$ feature only 7 and 4 modal operators, respectively; to ease the notation we use the original one, so that $DC$, for example, denotes the relation {\em disconnected} and $\langle DC\rangle$ its corresponding modal operator; examples are shown in Fig.~\ref{fig:dir-top} (right).
%

\smallskip

\noindent{\bf Some known result.} Although $\mathcal {HS}$$^k$ has not been studied per se, the literature that concerns its fragments is very wide. Very briefly, it is worth recalling that satisfiability for $\mathcal {HS}$, its one-dimensional version, is undecidable~\cite{HS}, and that various strategies have been considered in the literature to define fragments or variants of $\mathcal {HS}$ with better computational behaviour. These include constraining the underlying temporal structure~\cite{DBLP:conf/jelia/MontanariSV02}, restricting the set of modal operators~\cite{DBLP:journals/acta/AcetoMGIMS16,DBLP:journals/tcs/BresolinMMSS14}, softening the semantics to a reflexive one~\cite{DBLP:journals/fuin/MarcinkowskiM14,DBLP:conf/time/MontanariPS10}, restricting the nesting of modal operators~\cite{DBLP:journals/amai/BresolinMMS14}, restricting the propositional power of the languages~\cite{DBLP:journals/tocl/BresolinKMRSZ17}, and considering \emph{coarser} interval temporal logics based on interval relations that describe a less precise relationship between intervals (similarly to what topological relations do)~\cite{DBLP:journals/ai/Munoz-VelascoPS19}. In the case of $\mathcal {HS}$$^2$, only the sub-languages for $\mathcal {HS}$$^2_{RCC8}$ and $\mathcal {HS}$$^2_{RCC5}$ have been studied, in~\cite{LRCC8}, and their satisfiability problem is undecidable as well, even under very simple assumptions, or can be proven so by exploiting the results on the one-dimensional case. In general, one can expect deduction in $\mathcal {HS}$$^k$ to be
a computationally hard problem even under very restrictive assumptions, such as finite domains. Although in the spirit of the existing work for interval temporal logic one can imagine exploring fragments of $\mathcal {HS}$$^k$, here we are interested in induction problems, for which expressive power and the possibility of describing patterns are more important qualities. In the following, we shall focus on bidimensional images, and therefore fix $k=2$. The particular sub-language used in a given knowledge extraction problem is also a parameter, and we can choose, for example, to learn patterns in $\mathcal {HS}^2$, $\mathcal {HS}^2_{RCC8}$, or $\mathcal {HS}^2_{RCC5}$.

\begin{figure}[t]
		\centering
		\includegraphics[width=1.\linewidth]{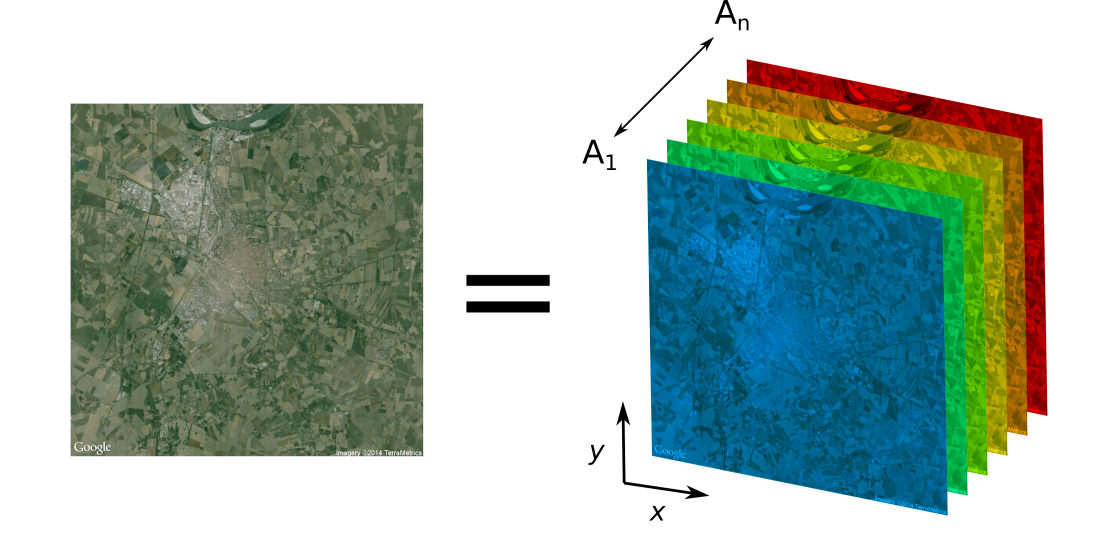}
		\caption{The structure of a multi-channel image. Satellite image courtesy of Google Earth.}
		\label{fig:img-struct}
\end{figure}

\section{Decision tree learning} \label{sec:dt-learning}

\noindent{\bf A theory of binary spatial decision trees.} In the bidimensional setting, a \emph{spatial dataset} is a set $\mathcal{S} = \{S_1,\ldots,S_m\}$ of $m$ spatial instances defined over a set of $n$ \emph{spatial} attributes $A_1,\ldots,A_n$; each attribute is a single-channel image with $N_S$ points $\pi_1,\ldots,\pi_{N_S}$, and we assume that such points form a bidimensional rectangular matrix, that is, $N_S=N_1\cdot N_2$ for some $N_1,N_2\in\mathbb N^+$, $N_1,N_2>1$. Thus
we fix $N=\max\{N_1,N_2\}$, $D=[1,\ldots,N]$, and $\mathbb D=\langle D,<\rangle$. In this way, we can interpret spatial datasets as models of $\mathcal {HS}$$^2$, in which each point $\pi$ is a pair $(x_1,x_2)\in\mathbb D^2$.
A \emph{categorical labelled} spatial dataset is a spatial dataset where the instances are associated to a \emph{target variable} $\mathcal{C} = \{C_1,\ldots,C_l\}$, known as \emph{class variable}. So, each image is described by many attributes, each of which can be thought of as a color channel, and expresses a real number for each spatial point (see Fig.~\ref{fig:img-struct}).
Digital images often depict objects in terms of colors from the visible spectrum, and usually encompass only a few color channels (e.g., three when using the popular RGB standard); however, there are situations where channels have a different interpretation (e.g., {\em hyperspectral imagery}). 
For a spatial attribute $A \in \mathcal{A}$, let $dom(A)$ denote its \emph{domain} (i.e., the set of values of $A$ that actually occur in $S$), and $A(\pi)$ the value of $A$ at the point $\pi$. The \emph{class} of an instance $S \in \mathcal{S}$ is denoted by $C(S)$. We assume that there are no missing values.
We shall describe properties of images by expressing properties of {\em rectangles},
 each denoted by a pair of points: $r=[(x_1,y_1),(x_2,y_2)]$. Such properties are called {\em propositional decisions}:
$$\mathcal W_{prop}=\{A\bowtie_\gamma a\},$$

\noindent where $\bowtie \, \in\{<,\le,=\,,\neq\}$, and are evaluated on rectangles. The parameter $\gamma\in (0,1]$ allows us avoid a too strict evaluation of such propositions on rectangles; in particular, we first define, for a point $\pi=(z,t)$:
$$(z,t)\in[(x_1,y_1),(x_2,y_2)] \stackrel{def}{\Leftrightarrow} x_1\le z<y_1\wedge x_2\le t<y_2,$$

\noindent and, then, we establish that $A\bowtie_\gamma a$ is true on a rectangle $r$ if and only if it holds:
$$\frac{|\{\pi\mid \pi\in r, A(\pi)\bowtie a\}|}{|\{\pi\mid \pi\in r\}|}\ge\gamma.$$

\noindent Fixed an image $S\in\mathcal S$, and a rectangle $r$ in it, we denote the above situation as: 
$$S,r\mmodels A\bowtie_\gamma a.$$

\noindent Our theory of spatial decision trees is based on $\mathcal {HS}^2$ (although, as we have recalled, the actual sub-language used can be a parameter of the problem), thus, on top of propositional decisions, the language of spatial decision trees encompasses a set of \emph{modal decisions}:
\[
\mathcal{W}_{mod} = \{\langle X_1,X_2 \rangle (A \bowtie_\gamma a)\},
\]

\noindent where $\langle X_1,X_2 \rangle$ is a $\mathcal {HS}^2$ modality.
Together, propositional and modal decisions form a set of {\em decisions}:
$$\mathcal W=\mathcal W_{prop}\cup\mathcal W_{mod}.$$

\noindent Binary \emph{spatial decision trees} are formulas of the following grammar:

\[
\tau ::=  (W_{prop} \land \tau) \lor (\neg W_{prop} \land \tau) \mid (W_{mod} \land \tau) \lor (\neg W_{mod} \land \tau ) \mid C
\]


\noindent where $W_{prop}\in\mathcal W_{prop}$ and $W_{mod}\in\mathcal W_{mod}$ are decisions and $C\in\mathcal C$ is a class. To define the semantics of spatial decision trees on a spatial dataset, we need to first define how decisions are evaluated. In the static case of binary decision trees, from a dataset one immediately computes its partition into the two subsets that are entailed by a propositional decision. In the spatial case, we start by assuming that each instance $S$ is {\em anchored} to a set of rectangles in the set $\mathbb K(\mathbb D^2)$, denoted $S.refs$. At the beginning of the learning phase, $S.refs=\{r_0\}$ for every $S$, where $r_0$ is a distinguished rectangle that we interpret as a privileged observation point from which the learning takes place; for example, it can be the central unit rectangle, or a rectangle at a corner of the image. Spatial decision tree learning is a {\em local} learning process; the local nature of decision trees does not transpire at the static level, but it becomes evident at the modal one. Every decision entails, potentially, new reference rectangles for every instance of a dataset. In particular, given an image $S$ with associated $S.refs$, and given a decision $W$, we can compute a set of {\em new reference rectangles} $f(S.refs,W)$ as:
$$\{s\in\mathbb{K}(\mathbb D^2) \mid \exists r\in S.refs \land r(R_{X_1},R_{X_2})s \land S,s \mmodels A \bowtie_\gamma a\}$$

\noindent if $W=\langle X_1,X_2 \rangle (A \bowtie_\gamma a)$, and as:
$$\{s\in S.refs \mid S,s \mmodels A \bowtie_\gamma a\}$$

\noindent if $W=A \bowtie_\gamma a$. When $W$ is clear from the context, we use $S.refs'$ to denote $f(S.refs,W)$. Splitting a dataset $\mathcal S$ on the basis of a decision becomes now possible. Given a spatial dataset $\mathcal S$ and a decision $W$, we say that
$$\mathcal{S}_e = \{ S \in \mathcal{S} \mid S.refs'\neq\emptyset \},$$

\noindent and we assume that each image $S\in\mathcal S_e$ is anchored to $S.refs'$, and
$$\mathcal{S}_u= \mathcal S\setminus\mathcal S_e,$$

\noindent whereas instances in $\mathcal S_u$ remain anchored to their original set of rectangles. For a decision $W$, we use the notation $S \mmodels W$  or $S, S.refs \mmodels W$ (respectively, $S \mmodels \neg W$ or $S, S.refs \mmodels \neg W$) to identify the members of $\mathcal{S}_e$ (respectively, $\mathcal{S}_u$). To define how a binary spatial decision tree is evaluated on a spatial dataset we need to establish how single instances are classified. Thus, we denote by $\tau(S,S.refs)$ the class assigned by $\tau$ to an instance $S$ anchored at the set $S.refs$, and define it inductively as:
\[
\begin{array}{lcl}
C & \text{if} & \tau=C, \\
\tau_1(S, S.refs') & \text{if} & \tau = (W \land \tau_1) \lor (\neg W \land \tau_2) \text{ and } S, S.refs \mmodels W, \\
\tau_2(S, S.refs) & \text{if} & \tau = (W \land \tau_1) \lor (\neg W \land \tau_2) \text{ and } S, S.refs \not\mmodels W. \\
\end{array}
\]

\noindent Then, we say that the class {\em assigned by} $\tau$ {\em to} $S$ is $\tau(S)=\tau(S,\{r_0\})$. Independently from the problem being propositional, temporal, or spatial, {\em evaluating} a tree $\tau$ on a dataset $\mathcal S$ entails classifying all its instances, and comparing the resulting class with the ground truth; such a comparison gives rise to a confusion matrix $\theta$, and, in turn, to define the notion:
$$\mathcal S\mmodels_\theta \tau.$$

\noindent When such an evaluation is performed on the training set, $\theta$ represents the {\em training performance} of $\tau$, while when it is performed on a classified dataset not used for training is called {\em test performance}.
%
%
%
%
%

\smallskip

\noindent{\bf Entropy based learning of spatial decision trees.} The decision tree model was originally defined by~\cite{DBLP:journals/ml/Quinlan86}, which resulted in the development of the \emph{Iterative Dichotomizer~3 (ID3)} algorithm, that is able to extract decision trees from static categorical data, and its extension \emph{C4.5}~\cite{Quinlan93}, that includes the possibility of dealing with numerical attributes as well. {\em Spatial C4.5} extends {\em C4.5} to deal with spatial datasets, preserving its driving principles. Since learning optimal trees is a NP-hard problem at the propositional level already~\cite{DBLP:journals/ipl/HyafilR76}, so is at the spatial level, and {\em Spatial C4.5}, just like {\em C4.5}, implements a suboptimal, greedy approach generally known as {\em entropy-based learning}. Let $\xi_i$ be the fraction of instances labelled with class $C_i$ in a dataset $\mathcal{S}$ with $\ell$ distinct classes. Then, the \emph{information conveyed} by $\mathcal{S}$ (or \emph{entropy} of $\mathcal{S}$) is computed as:

\[
Info(\mathcal{S}) = - \sum_{i=1}^\ell \xi_i \mathrm{log} \xi_i.
\]

\noindent Intuitively, the entropy is inversely proportional to the purity degree of $\mathcal{S}$ with respect to the class values. In binary trees, {\em splitting}, which is the main greedy operation, is performed over a specific attribute $A$, a threshold value $a\in dom(A)$, a value $\gamma$, and an operator $\bowtie$. Let $W(A,a,\gamma,\bowtie)$ be the decision entailed by $A,a,\gamma,\bowtie$, and let $(\mathcal S_e,\mathcal S_u)$ be the partition of $\mathcal S$ entailed by $W(A,a,\gamma,\bowtie)$ (as defined above). The {\em splitting information} of $W=W(A,a,\gamma,\bowtie)$ is defined as:

\[
InfoSplit(\mathcal{S},W) = \frac{|\mathcal S_e|}{|\mathcal S|}Info(\mathcal S_e)+\frac{|\mathcal S_u|}{|\mathcal S|}Info(\mathcal S_u).
\]

\noindent In this way, we can define the {\em entropy gain of a decision} as:

\[
InfoGain(\mathcal{S},W) = Info(\mathcal S)-InfoSplit(\mathcal S,W).
\]

\begin{algorithm}[t!]
\SetKwInOut{Input}{input}\SetKwInOut{Output}{output}
\SetKwProg{Function}{function}{:}{end function}
\SetKwComment{Comment}{$\triangleleft$\ }{}
\DontPrintSemicolon
\Input{$\mathcal{S}$ - spatial dataset}
\Output{$\tau$ - spatial decision tree}
\Function{$SpatialC4.5(\mathcal{S})$}{
	$Preprocess(\mathcal{S})$\\
	$\tau \gets Learn(\mathcal{S})$\\
	\Return{$\tau$}
}
\Input{$\mathcal{S}$ - spatial dataset}
\Output{$\nu$ - decision node}
\Function{$Learn(\mathcal{S})$}{
	\lIf{some stopping condition applies}{\Return{$CreateLeafNode(\mathcal{S})$}}
	$W \gets FindBestDecision(\mathcal{S})$ \\
	$( \mathcal{S}_e, \mathcal{S}_u ) \gets Split(\mathcal{S},W)$ \\
	$\nu \gets CreateNode(\mathcal{S},W)$ \\
	$\nu.left \gets Learn(\mathcal{S}_e)$ \\
	$\nu.right \gets Learn(\mathcal{S}_u)$ \\
	\Return{$\nu$}
}
\caption{High-level description of {\em Spatial C4.5}.}
\label{alg:c4.5}
\end{algorithm}


\noindent Given a spatial dataset $\mathcal S$, the language of all possible decisions, and therefore of all possible splits, is implicitly defined, and the generic, high level {\em Spatial C4.5} algorithm works as shown in Alg.~\ref{alg:c4.5}. This learning algorithm returns a binary tree $\tau$, in which every node is associated with a subset of $\mathcal S$, and every leaf node is associated with a class that depends on the labeling function that has been used. The most typical choice is labeling a leaf with the class that occurs the most in the associated dataset, but every labeling function, in general, generates a certain amount of misclassification.
As in the propositional case, the procedure includes some stopping criteria, preventing the resulting model from overfitting to the training data.
The stopping conditions, also called {\em pre-pruning} conditions, depend on the particular implementation and are subject to statistical evaluations; typical conditions include testing the purity of the dataset associated with the candidate-leaf node, evaluating its cardinality, and evaluating the information gain that may emerge from a further splitting step.
Note, however, that there are other approaches to prevent overfitting (e.g., training {\em ensambles} of trees).

The theoretical complexity of this algorithm mainly depends on the number of possible decisions, as well as the cost of checking the truth of each decision on each instance. Both at the propositional and spatial level, the time complexity is polynomial, but the truth-checking step in the spatial case adds a $\Theta (N^{2k})$ term (note that this is the number of different hyperrectangles in an $N \times \ldots \times N$, $k$-dimensional dimensional space) that is not present at the propositional level.
The parameters of {\em Spatial C4.5} includes all classical propositional parameters (e.g., the stopping conditions) plus the subset of potential decisions, which in turn depends on the subset of (direct or disjunctive) operators of $\mathcal {HS}^2$ and the admitted values of $\gamma$, and the location of the starting hyperrectangle $r_0$.

\smallskip

\noindent{\bf \emph{Spatial C4.5} implementation.} Implementations of decision tree learning algorithms exist in several learning suites; the most popular implementations are provided by the Scikit-learn suite~\cite{scikit-learn} (in Python), and Weka~\cite{weka} (in Java).
In recent years, the Julia programming language is becoming increasingly popular for scientific computing and, although the language is still young, there exists a stable Julia package for decision tree learning~\cite{BensadeghiJuliaDecisionTreeModule}.
Due to the performance gains that Julia, as a compiled language, enables, we developed an implementation of {\em Spatial C4.5} starting from the existing Julia package.

\section{Experimental results} \label{sec:exp-results}

\noindent{\bf Land cover classification.} While spatial decision tree learning can be applied to many image-related tasks,
we focus here on a single image classification problem that is suitable for small-scale, spatial, raw-pixel reasoning:
{\em land cover classification}. 
The objective is to label each pixel in aerial/satellite imagery with
a class label carrying meaningful information about its content (e.g., classify it as {\em forest}, {\em urban area}, or {\em crop field}).
Due to the high altitude from which the images are captured, a single pixel carries the average electromagnetic response of a wider surface area (e.g., one to hundreds of squared meters) and, as such, the classification of a pixel in the image only depends on a close neighborhood of pixels. 
One can see how learning spatial patterns can be computationally easy in this context. 
Strictly speaking, land cover classification is an instance of {\em image segmentation} but,
because of its properties, the problem has been often tackled by classifying each pixel independently. 
This choice is also supported by the fact that the imagery involved is often hyperspectral, that is, composed of many color channels, describing the strength of signals at different electromagnetic wavelengths; thus, each pixel holds a large number of values (e.g., usually in the hundreds) that, collectively, describe in a satisfactory way the nature of its content.
Within this framework, a pixel can be represented as a single line of tabular data, and propositional symbolic methods naturally lend themselves for the job; together with the need for explicit classification rules and good performances, this is one of the reasons why this task has a long history being approached by propositional symbolic methods~\cite{WorldView2,DTRBRF,Corn,KulkarniImageAnalysis,DBLP:journals/entropy/XuLWCZX21,DBLP:journals/remotesensing/YangWDSLW17,InferringMethodLand}. 
Although it proved fruitful in the past, classifying a pixel solely according to its spectral footprint is limited, as it disregards the spatial structure of the image. In recent times there have been many neural network-based applications that can capture spatial, inter-pixel dependencies in some sense~\cite{DBLP:journals/corr/abs-1904-10674,DBLP:journals/corr/CaoZXMXP17,DBLP:journals/lgrs/LiangZGZ20,DBLP:journals/lgrs/RoyKDC20,DBLP:journals/tgrs/SantaraMHSGPM17,DBLP:journals/access/ZhangWZXLCB20}.
These approaches leverage intrinsic spatial localities through the engineering of spatial features, and their performances are ultimately better than those of previous attempts; however, as we have already observed, as a general rule functional approaches give up the interpretability and the explainability of the resulting models.

\smallskip

\noindent{\bf Experimental setup.} We have identified four datasets that are publicly available and commonly used to benchmark neural network-based methods for hyperspectral land cover classification.
They are referred to as {\em Indian Pines}, {\em Pavia University}, {\em Pavia Centre}, and {\em Salinas-A}. Each dataset consists of an image of a piece of land, coupled with a {\em ground-truth label mask}, providing the correct class for some of the pixels in the image. In all cases, the image is captured by a dedicated sensor during a flight campaign: specifically, {\em Pavia University} and {\em Pavia Centre} were collected using a ROSIS sensor (Reflective Optics System Imaging Spectrometer) and the remaining ones using an AVIRIS sensor (Airborne Visible/Infrared Imaging Spectrometer). The ROSIS and AVIRIS yield spectral detections covering a range of frequencies from $0.43 \mu m$ to $0.86 \mu m$ and from $0.4 \mu m$ to $2.45 \mu m$, with a number of channels of $103$ and $200$, respectively. The size of the images varies from $86   \times 83  $ pixels for the {\em Salinas-A} dataset to $1096 \times 1096$ pixels for {\em Pavia Centre}; however, not all pixels are labelled. 
The typical approach to using these datasets for land cover classification involves gathering either all labelled pixels, or a randomly sampled subset, and applying a classifier model extraction to them. Some approaches also make use of the spatial structure, and consider, for the classification of each pixel, a set of neighboring pixels; the neighborhood is generally in the form of a $d\times d$ window centered in the pixel, for a natural odd number $d$. This one is also our solution, so that each dataset, for us, is a collection of $d\times d$ images. In general, the chosen datasets present somewhat skewed raw data; to correct the biases towards certain classes, in this experiment the class counts are balanced according to a simple transformation: a fixed number $P$ of pixels is randomly sampled for each class, discarding less numerous classes. 
In this experiment, we fixed $d=3$ and $P=100$; this entailed producing the following classification sub-datasets: {\em Indian Pines} with 12 classes (and therefore 1200 instances), {\em Pavia University} with 9 (900), {\em Pavia Centre} with 9 (900), and {\em Salinas-A} with 6 (600).
We used three of these four datasets for a systematic comparisons between different approaches to symbolic classification, and the fourth one to specifically extract interpretable models and show their potential. Each dataset of the first group ({\em Indian Pines}, {\em Pavia University}, and {\em Pavia Centre}) is then split in two balanced sets for training (80\%) and test (20\%); each execution, with a different seed (from 1 to 10), entails a different training/test pair. We compared the performances of the following approaches: \begin{inparaenum}[\it (i)] \item {\em Spatial C4.5} with $\mathcal {HS}^2_{RCC5}$ (simply referred to as $\mathcal {HS}^2_{RCC5}$); \item {\em Spatial C4.5} with $\mathcal {HS}^2_{RCC8}$ ($\mathcal {HS}^2_{RCC5}$); \item {\em propositional single-pixel} ({\em single-pixel}), in which we use standard C4.5 trees learned on the numerical description of the pixel to be classified, without spatial information; \item {\em propositional flattened} ({\em flattened}), in which we use standard C4.5 trees learned on the numerical descriptions of all pixels in the $d\times d$ image, without spatial information, and \item {\em propositional averaged} ({\em averaged}), in which we use standard C4.5 trees learned on single pixels described by averaging all values of the $d\times d$ image, again, without spatial information. \end{inparaenum} The baseline is clearly the simple approach, that classifies a single pixel based on its own spectral footprint (for example in the case of {\em Indian Pines}, this implies that a single pixel is classified using the values of the 200 spectral attributes describing it, but no values of any neighboring pixel). The flattened approach is the spatial equivalent of using {\em lagged values} in standard temporal learning; in this case, the information of the neighboring pixel is included, but the spatial information is not preserved (again, in the example of {\em Indian Pines}, this means that each pixel is classified according to $d^2 \times 200$, unordered attributes): as we shall see, this method performs generally bad, probably because of a too large number of variables, which has a negative effect on entropy-based, locally optimal learning. The averaged approach is based on describing every pixel with the average value of all its neighboring ones; in this way the spatial component is included in the learning, but in a unordered way: unlike the flattened version, however, the number of variables is again equal to the number of channels, allowing standard decision trees to work well enough. All approaches share the same implementation of {\em Spatial C4.5}; in the propositional experiments, the modal part is simply excluded.
With regards to the pre-pruning parametrization, after some preliminary tests, three conditions were fixed: a minimum number of samples per leaf to $4$, a minimum information gain of $0.01$ for a split to be meaningful, and a maximum entropy at each leaf of $0.3$. As for the spatial parametrization, we fixed $r_0$ to be always the minimum rectangle that contains the pixel to be classified. Moreover, to define the set of allowed decisions, we fixed $\bowtie\ =\{\le,\ge\}$ and $\gamma\in\{0.6,0.7,0.8,0.9,1\}$. Finally, in terms of performance metrics, we adopted: \begin{inparaenum}[\it (i)]\item {\em accuracy}, \item {\em kappa coefficient} (which relativizes the accuracy to the probability of a random correct answer), \item {\em sensitivity}, \item {\em specificity}, and \item {\em precision} \end{inparaenum} all computed in test mode.
Each experiment is repeated a number times with different seeds for the sampling steps; this number was fixed to $10$ to satisfy some time constraints.

\smallskip

\afterpage{\clearpage}
\begin{sidewaysfigure}[h!]
\scriptsize
\centering
\setlength{\tabcolsep}{1pt}
\renewcommand\arraystretch{1.37} 
 \resizebox{0.9\textheight}{!}{
\begin{tabular}{c|c*{5}{||lllll}}
\hline
\multirow{2}{*}{dataset} & \multirow{2}{*}{run} &
	\multicolumn{5}{c}{single-pixel} &
	\multicolumn{5}{c}{flattened} &
	\multicolumn{5}{c}{average} &
	\multicolumn{5}{c}{$\mathcal {HS}^2_{RCC8}$} &
	\multicolumn{5}{c}{$\mathcal {HS}^2_{RCC5}$} \\
&
& $\kappa$ & acc & sens & spec & prec
& $\kappa$ & acc & sens & spec & prec
& $\kappa$ & acc & sens & spec & prec
& $\kappa$ & acc & sens & spec & prec
& $\kappa$ & acc & sens & spec & prec
\\
\hline
\hline
\parbox[t]{2mm}{\multirow{10}{*}{\rotatebox[origin=c]{-90}{\emph{Indian Pines}}}}
& 1  &  69.09  &  71.67  &  45  &  98.64  &  75
     &  51.82  &  55.83  &  75  &  97.73  &  75
     &  69.55  &  72.08  &  75  &  97.73  &  75
     &  70.91  &  73.33  &  95  &  97.27  &  76
     &  73.18  &  75.42  &  65  &  98.18  &  76.47
     \\
& 2  &  55.45  &  59.17  &  45  &  95.91  &  50
     &  54.09  &  57.92  &  55  &  98.64  &  78.57
     &  70  &  72.5  &  75  &  97.73  &  75
     &  70.91  &  73.33  &  80  &  97.27  &  72.73
     &  74.09  &  76.25  &  75  &  96.36  &  65.22
     \\
& 3  &  66.82  &  69.58  &  75  &  98.64  &  83.33
     &  63.64  &  66.67  &  75  &  96.82  &  68.18
     &  62.27  &  65.42  &  55  &  97.27  &  64.71
     &  72.73  &  75  &  55  &  98.64  &  78.57
     &  68.64  &  71.25  &  70  &  98.18  &  77.78
     \\
& 4  &  58.64  &  62.08  &  65  &  96.36  &  61.9
     &  60.45  &  63.75  &  55  &  98.18  &  73.33
     &  70  &  72.5  &  75  &  97.27  &  71.43
     &  72.27  &  74.58  &  75  &  98.18  &  78.95
     &  67.27  &  70  &  55  &  99.09  &  84.62
     \\
& 5  &  65.45  &  68.33  &  70  &  97.73  &  73.68
     &  55.45  &  59.17  &  60  &  98.64  &  80
     &  67.27  &  70  &  70  &  97.73  &  73.68
     &  71.82  &  74.17  &  75  &  100  &  100
     &  71.82  &  74.17  &  85  &  98.18  &  80.95
     \\
& 6  &  58.18  &  61.67  &  50  &  96.82  &  58.82
     &  54.55  &  58.33  &  60  &  96.36  &  60
     &  64.55  &  67.5  &  90  &  96.82  &  72
     &  68.64  &  71.25  &  70  &  98.18  &  77.78
     &  72.73  &  75  &  90  &  98.64  &  85.71
     \\
& 7  &  63.64  &  66.67  &  40  &  97.27  &  57.14
     &  57.27  &  60.83  &  85  &  96.36  &  68
     &  66.82  &  69.58  &  80  &  99.55  &  94.12
     &  70.91  &  73.33  &  70  &  99.09  &  87.5
     &  71.36  &  73.75  &  65  &  98.64  &  81.25
     \\
& 8  &  58.18  &  61.67  &  70  &  98.64  &  82.35
     &  59.55  &  62.92  &  55  &  96.36  &  57.89
     &  68.18  &  70.83  &  70  &  96.36  &  63.64
     &  66.36  &  69.17  &  75  &  96.82  &  68.18
     &  67.73  &  70.42  &  75  &  96.82  &  68.18
     \\
& 9  &  68.64  &  71.25  &  65  &  99.09  &  86.67
     &  52.73  &  56.67  &  85  &  96.82  &  70.83
     &  71.82  &  74.17  &  80  &  98.18  &  80
     &  77.27  &  79.17  &  85  &  98.18  &  80.95
     &  71.36  &  73.75  &  80  &  98.64  &  84.21
     \\
& 10  &  60  &  63.33  &  60  &  98.64  &  80
     &  55.45  &  59.17  &  70  &  96.82  &  66.67
     &  71.82  &  74.17  &  85  &  96.36  &  68
     &  76.36  &  78.33  &  100  &  99.09  &  90.91
     &  76.36  &  78.33  &  75  &  99.09  &  88.24
     \\
\hline
&avg  &  62.41  &  65.54  &  58.50  &  97.77  &  70.89
     &  56.50  &  60.13  &  67.50  &  97.27  &  69.85
     &  68.23  &  70.88  &  75.50  &  97.50  &  73.76
     &  71.82  &  74.17  &  78.00  &  98.27  &  81.16
     &  71.45  &  73.83  &  73.50  &  98.18  &  79.26
     	\\
\hline
\hline
\parbox[t]{2mm}{\multirow{10}{*}{\rotatebox[origin=c]{-90}{\emph{Pavia University}}}}
& 1  &  74.38  &  77.22  &  85  &  96.25  &  73.91
     &  39.38  &  46.11  &  45  &  93.12  &  45
     &  78.75  &  81.11  &  70  &  97.5  &  77.78
     &  81.25  &  83.33  &  90  &  96.25  &  75
     &  80  &  82.22  &  90  &  96.25  &  75
     \\
& 2  &  74.38  &  77.22  &  75  &  96.88  &  75
     &  36.25  &  43.33  &  60  &  93.75  &  54.55
     &  80.62  &  82.78  &  75  &  98.12  &  83.33
     &  83.12  &  85  &  80  &  97.5  &  80
     &  80.62  &  82.78  &  75  &  97.5  &  78.95
     \\
& 3  &  74.38  &  77.22  &  70  &  95  &  63.64
     &  50  &  55.56  &  75  &  96.25  &  71.43
     &  82.5  &  84.44  &  80  &  98.12  &  84.21
     &  90.63  &  91.67  &  100  &  100  &  100
     &  86.25  &  87.78  &  100  &  99.38  &  95.24
     \\
& 4  &  67.5  &  71.11  &  75  &  95.62  &  68.18
     &  39.38  &  46.11  &  50  &  93.75  &  50
     &  74.38  &  77.22  &  80  &  95.62  &  69.57
     &  84.38  &  86.11  &  90  &  98.75  &  90
     &  84.38  &  86.11  &  85  &  99.38  &  94.44
     \\
& 5  &  73.12  &  76.11  &  75  &  96.88  &  75
     &  44.38  &  50.56  &  50  &  94.38  &  52.63
     &  80  &  82.22  &  75  &  98.75  &  88.24
     &  78.12  &  80.56  &  75  &  95  &  65.22
     &  81.88  &  83.89  &  80  &  99.38  &  94.12
     \\
& 6  &  76.88  &  79.44  &  65  &  99.38  &  92.86
     &  32.5  &  40  &  25  &  90.62  &  25
     &  79.38  &  81.67  &  85  &  95.62  &  70.83
     &  75  &  77.78  &  75  &  98.75  &  88.24
     &  78.12  &  80.56  &  90  &  98.75  &  90
     \\
& 7  &  73.12  &  76.11  &  70  &  97.5  &  77.78
     &  41.25  &  47.78  &  50  &  95.62  &  58.82
     &  80.62  &  82.78  &  70  &  98.12  &  82.35
     &  82.5  &  84.44  &  85  &  96.88  &  77.27
     &  78.75  &  81.11  &  85  &  96.88  &  77.27
     \\
& 8  &  70.62  &  73.89  &  60  &  96.88  &  70.59
     &  40  &  46.67  &  40  &  90.62  &  34.78
     &  71.25  &  74.44  &  65  &  96.88  &  72.22
     &  82.5  &  84.44  &  90  &  97.5  &  81.82
     &  83.75  &  85.56  &  95  &  96.25  &  76
     \\
& 9  &  73.12  &  76.11  &  75  &  98.12  &  83.33
     &  38.12  &  45  &  50  &  90.62  &  40
     &  75  &  77.78  &  65  &  96.88  &  72.22
     &  80.62  &  82.78  &  90  &  98.75  &  90
     &  78.12  &  80.56  &  90  &  98.75  &  90
     \\
& 10  &  74.38  &  77.22  &  80  &  98.12  &  84.21
     &  38.12  &  45  &  30  &  94.38  &  40
     &  74.38  &  77.22  &  55  &  96.88  &  68.75
     &  86.25  &  87.78  &  95  &  98.75  &  90.48
     &  86.87  &  88.33  &  95  &  98.75  &  90.48
     \\
\hline
&avg  &  73.19  &  76.17  &  73.00 &  97.06  &  76.45
     &  39.94  &  46.61  &  47.50  &  93.31  &  47.22
     &  77.69  &  80.17  &  72.00  &  97.25  &  76.95
     &  82.44  &  84.39  &  87.00  &  97.81  &  83.80
     &  81.87  &  83.89  &  88.50  &  98.13  &  86.15
          \\
\hline
\hline
\parbox[t]{2mm}{\multirow{10}{*}{\rotatebox[origin=c]{-90}{\emph{Pavia Centre}}}}
& 1  &  85  &  86.67  &  90  &  98.12  &  85.71
     &  34.38  &  41.67  &  45  &  92.5  &  42.86
     &  86.87  &  88.33  &  90  &  96.25  &  75
     &  88.75  &  90  &  80  &  97.5  &  80
     &  88.75  &  90  &  80  &  97.5  &  80
     \\
& 2  &  83.12  &  85  &  90  &  98.12  &  85.71
     &  42.5  &  48.89  &  40  &  96.88  &  61.54
     &  90.63  &  91.67  &  90  &  98.12  &  85.71
     &  91.25  &  92.22  &  90  &  98.75  &  90
     &  89.38  &  90.56  &  90  &  98.75  &  90
     \\
& 3  &  83.12  &  85  &  70  &  99.38  &  93.33
     &  40.62  &  47.22  &  55  &  96.25  &  64.71
     &  85.62  &  87.22  &  90  &  96.88  &  78.26
     &  86.25  &  87.78  &  95  &  98.12  &  86.36
     &  86.25  &  87.78  &  95  &  98.12  &  86.36
     \\
& 4  &  84.38  &  86.11  &  80  &  97.5  &  80
     &  45.63  &  51.67  &  75  &  91.88  &  53.57
     &  91.25  &  92.22  &  90  &  98.12  &  85.71
     &  91.88  &  92.78  &  95  &  98.12  &  86.36
     &  91.88  &  92.78  &  85  &  98.12  &  85
     \\
& 5  &  85.62  &  87.22  &  90  &  96.88  &  78.26
     &  30  &  37.78  &  40  &  95  &  50
     &  90.63  &  91.67  &  80  &  97.5  &  80
     &  90  &  91.11  &  80  &  100  &  100
     &  87.5  &  88.89  &  85  &  98.12  &  85
     \\
& 6  &  83.12  &  85  &  90  &  99.38  &  94.74
     &  38.12  &  45  &  50  &  93.75  &  50
     &  86.87  &  88.33  &  90  &  98.12  &  85.71
     &  85  &  86.67  &  95  &  96.25  &  76
     &  85  &  86.67  &  95  &  96.25  &  76
     \\
& 7  &  84.38  &  86.11  &  90  &  98.75  &  90
     &  43.75  &  50  &  55  &  95  &  57.89
     &  90  &  91.11  &  95  &  98.12  &  86.36
     &  91.25  &  92.22  &  85  &  98.75  &  89.47
     &  91.25  &  92.22  &  90  &  98.12  &  85.71
     \\
& 8  &  83.75  &  85.56  &  80  &  97.5  &  80
     &  36.88  &  43.89  &  45  &  92.5  &  42.86
     &  93.12  &  93.89  &  80  &  98.75  &  88.89
     &  90.63  &  91.67  &  85  &  98.75  &  89.47
     &  83.12  &  85  &  70  &  97.5  &  77.78
     \\
& 9  &  89.38  &  90.56  &  90  &  97.5  &  81.82
     &  35  &  42.22  &  15  &  93.12  &  21.43
     &  87.5  &  88.89  &  85  &  98.75  &  89.47
     &  87.5  &  88.89  &  90  &  98.75  &  90
     &  90  &  91.11  &  100  &  98.75  &  90.91
     \\
& 10  &  83.75  &  85.56  &  90  &  99.38  &  94.74
     &  40.62  &  47.22  &  40  &  96.88  &  61.54
     &  93.75  &  94.44  &  100  &  98.75  &  90.91
     &  92.5  &  93.33  &  100  &  98.75  &  90.91
     &  91.25  &  92.22  &  95  &  98.75  &  90.48
     \\
\hline
&avg  &  84.56  &  86.28  &  86.00  &  98.25  &  86.43
     &  38.75  &  45.56  &  46.00  &  94.38  &  50.64
     &  89.62  &  90.78  &  89.00  &  97.94  &  84.60
     &  89.50  &  90.67  &  89.50  &  98.37  &  87.86
     &  88.44  &  89.72  &  88.50  &  98.00  &  84.72
          \\
\hline
\hline
\end{tabular}
 }
\caption{Comparison of the results on 3 datasets for land cover classification using propositional and spatial decision trees. Metrics, from left to right: kappa coefficient, accuracy, sensitivity, specificity, precision. All the results are in percentage points.}\label{tab:results}
\end{sidewaysfigure}

\begin{figure}[t]
\centering
\includegraphics[scale=0.8]{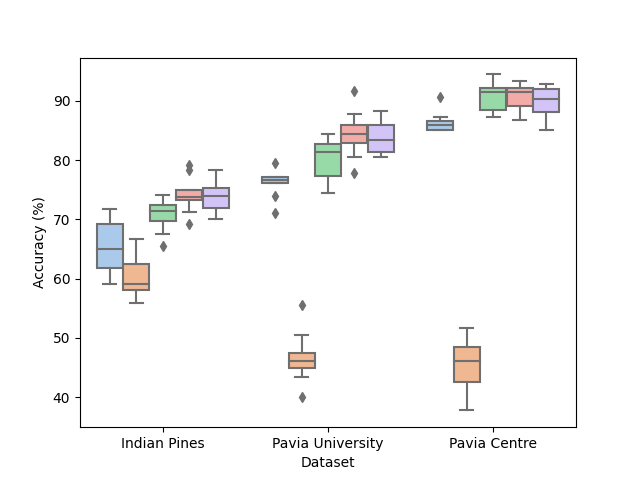}
\caption{Boxplot of the populations of accuracies produced by the 10 runs, per each dataset. In each group, from left to right, the approaches are ordered as follows: {\em simple}, {\em flattened}, {\em averaged}, $\mathcal {HS}^2_{RCC8}$, and $\mathcal {HS}^2_{RCC5}$.}\label{fig:box}
\end{figure}

\begin{figure*}[t]
\centering\footnotesize
\noindent\begin{minipage}[c]{.49\textwidth}
\resizebox{\textwidth}{!}{\begin{tikzpicture}
\node (0) at (0,0) [above] {$\tau$};
\node (00) at (-3,-2) {$\nu_1$};
\node (01) at (3,-2) {$\nu_2$};
\node (000) at (-5,-4) {$\nu_3$};
\node (001) at (-1,-4) {$C_1$};
\node (010) at (1,-4) {$\nu_4$};
\node (011) at (5,-4) {$C_4$};
\node (0000) at (-7,-6) {$C_6$};
\node (0001) at (-3,-6) {$C_5$};
\node[color=blue] (0100) at (-1,-6) {$C_2$};
\node (0101) at (3,-6) {$C_3$};
\path(0) edge[sloped,above] node {$A_{42} \geq 3367$} (00);
\path[color=blue](0) edge[sloped,above] node {$A_{42} < 3367$} (01);
\path(00) edge[sloped,above] node {$A_{14} \geq 1461$} (000);
\path(00) edge[sloped,above] node {$A_{14} < 1461$} (001);
\path[color=blue](01) edge[sloped,above] node {$A_{85} \geq 1296$} (010);
\path(01) edge[sloped,above] node {$A_{85} < 1296$} (011);
\path(000) edge[sloped,above] node {$A_{111} \leq 125$} (0000);
\path(000) edge[sloped,above] node {$A_{111} > 125$} (0001);
\path[color=blue](010) edge[sloped,above] node {$A_{7} \leq 1628$} (0100);
\path(010) edge[sloped,above] node {$A_{7} > 1628$} (0101);
\end{tikzpicture}}
{\scriptsize $$ A_{42} < 3367 \ \land \  A_{85} \geq 1296 \ \land \  A_{7} \leq 1628 \Rightarrow C_2$$}
\end{minipage}
\hfill
\noindent\begin{minipage}[c]{.49\textwidth}
\resizebox{\textwidth}{!}{\begin{tikzpicture}
\node (0) at (0,0) [above] {$\tau$};
\node (00) at (-3,-2) {$\nu_1$};
\node (01) at (3,-2) {$\nu_2$};
\node (000) at (-5,-4) {$\nu_3$};
\node (001) at (-1,-4) {$C_1$};
\node (010) at (1,-4) {$\nu_4$};
\node (011) at (5,-4) {$C_4$};
\node (0000) at (-7,-6) {$C_6$};
\node (0001) at (-3,-6) {$C_5$};
\node[color=blue] (0100) at (-1,-6) {$C_2$};
\node (0101) at (3,-6) {$C_3$};

\path(0) edge[sloped,above] node {$\langle \overline{NTPP}\rangle (A_{43} \geq_{0.8} 1978)$} (00);
\path[color=blue](0) edge[sloped,above] node {$[ \overline{NTPP}] (A_{43} <_{0.2} 1978)$} (01);
\path(00) edge[sloped,above] node {$A_{17} \geq 1743$} (000);
\path(00) edge[sloped,above] node {$A_{17} < 1743$} (001);
\path[color=blue](01) edge[sloped,above] node {$\langle EC\rangle (A_{83} \geq 638)$} (010);
\path(01) edge[sloped,above] node {$[EC] (A_{83} < 638)$} (011);
\path(000) edge[sloped,above] node {$A_{67} \geq_{0.6} 2656$} (0000);
\path(000) edge[sloped,above] node {$A_{67} <_{0.4} 2656$} (0001);
\path[color=blue](010) edge[sloped,above] node {$A_{6} \leq 1625$} (0100);
\path(010) edge[sloped,above] node {$A_{6} > 1625$} (0101);
\end{tikzpicture}}
{\scriptsize $$[ \overline{NTPP}] (A_{43} <_{0.2} 1978) \ \land\  \langle EC\rangle (A_{83} \geq 638 \ \land\  A_{6} \leq 1625) \Rightarrow C_2$$}
\end{minipage}
\caption{Comparison between a propositional (left) and a spatial (right) decision tree extracted on {\em Salinas-A}. The highlighted paths show the rules that each model has extracted for the class $C_2$ (i.e., {\em Corn senesced green weeds}). In this picture, $\bowtie_\gamma$ is denoted simply as $\bowtie$ when $\gamma=1.0$}
\label{fig:example-decision-trees}
\end{figure*}
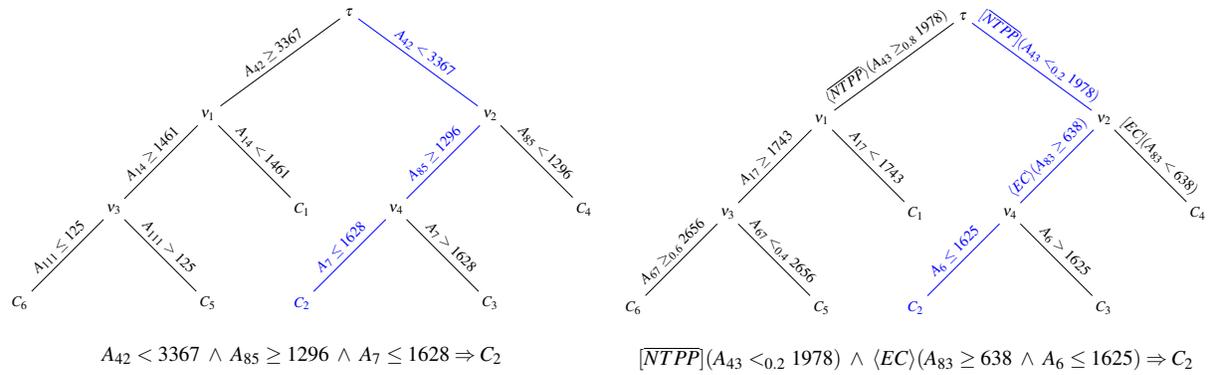

\noindent{\bf Results and discussion} Tab.~\ref{tab:results} shows a comparison of the results of the four different approaches in the three datasets. Overall, the total computation time needed for the experiments was around 96 hours on a Dell PowerEdge server with Intel Xeon Gold 6238R processors, using multi-threading (4 threads). A great portion of this time was spent in searching the locally optimal decision in the spatial approaches, which takes $10$ to $100$ times more time, compared to the propositional approaches.
Accuracy and $\kappa$ coefficient show similar trends, and only differ in a significant way when the overall performance is low. The lowest performance is achieved on {\em Pavia University} and {\em Pavia Centre} with the flattened approach; these are cases where no split satisfies the constraints (e.g., no split is informative enough), and the trees only consist of a leaf node, eventually classifying all samples with the same class. The accuracy in these cases is similar to that of a random classifier. At a closer look, it appears that the flattened approach is never competitive with the other methods, and is never better than propositional ones (incidentally, this is exactly the opposite to what happens using lagged values in temporal reasoning). Altogether, the results across the five datasets show that both spatial approaches always yield better results than non-spatial ones. The improvements seem to be proportional to the intrinsic hardness of each classification problem; compared with the propositional approach, the average improvement of {\em Spatial C4.5} ranges from 5\% to 9\%. The averaged approach performs better than the propositional one, but worse than the spatial one, except one case, {\em Pavia Centre}, in which the two approaches have essentially the same performance; yet, the averaged approach works with a computed feature (the average), so the original dataset and the averaged one are not really comparable to each other. Considering the test accuracies in the 10 executions per each dataset and per each approach, and interpreting them as populations, we also produced the boxplot for each dataset, in order to have a graphical understanding of the overall performances (see Fig.~\ref{fig:box}). One can observe how in two out of three cases, the spatial approaches behave better than all others. In two cases, the simple approach is very consistent, showing a very low variability ({\em Pavia University} and {\em Pavia Centre}), while this is not the case for the third one. Among the spatial approaches, $\mathcal {HS}^2_{RCC5}$ is always slightly worse than $\mathcal {HS}^2_{RCC8}$, and it has always higher variability; moreover, $\mathcal {HS}^2_{RCC8}$ presents no outliers. On top of the statistical analysis on the performances, we can see in Fig.~\ref{fig:example-decision-trees} (top) a pictorial example of two decision trees; in particular, these are trees extracted executing {\em Spatial C4.5} on a randomized training set extracted from {\em Salinas-A}. In this case, the problem to be solved is learning how to distinguish different types of vegetables. The one shown on the left side has been extracted by the propositional single-pixel approach, while the one on the right side has been extracted by the spatial RCC8 approach. The propositional tree presents $\kappa=92\%$, but a closer look reveals that the majority of the misclassifications concern the class $C_2$ ({\em corn senesced green weeds}), often confused with $C_6$ ({\em lettuce romaine, 7 weeks}) by this model (in particular, true $C_2$ instances are often classified as $C_6$), and the class-sensitivity of $C_2$ is only $75\%$. On the other hand, the spatial tree presents $\kappa=100\%$, which, in particular, implies that the class-sensitivity of all classes, and thus $C_2$ as well, is $100\%$. Since in this particular case both models have only one leaf per class, one can easily extract a rule that, in some sense, describes the spatial properties of this class. The two formulas that emerge are shown in Fig.~\ref{fig:example-decision-trees} (bottom). As we can see, the propositional one gives very limited information, that only concerns the attributes of the pixel to be classified. The spatial one (in this case, a $\mathcal{HS}^2_{RCC8}$-formula), instead, describes a richer spatial scenario, which could be translated in natural language as follows: {\em if the $3\times 3$ window surrounding the to-be-classified pixel presents strictly more than 20\% of pixels with $A_{43}<1978$, and at there exists at least one rectangle of neighboring pixels in which all of the pixels have $A_{83} \geq 638$ and $A_6 \leq 1625$, then the pixel is corn senesced green weeds}. In turn, such a formula could be translated to a spatial model, which, in essence, visually represents the class $C_2$. Thus, we can explain why the propositional tree fails, when it does: the first split is performed on the characteristics of the pixel itself, while the class $C_2$, seemingly, requires considerations on the windows in which it is submerged, to be correctly distinguished from $C_6$. This example shows the advantages of symbolic models over non-symbolic ones: the models can be read in natural language.

\section{Conclusion and future steps}

In this paper, we proposed a novel technique for knowledge extraction from spatial data. In the context of a general approach called {\em modal symbolic learning}, we considered a well-known symbolic learning schema, namely decision trees, and enhanced it by substituting propositional logic with a suitable modal spatial logic. With this method we can exploit the spatial arrangements of objects, and therefore make more informed decisions towards the classification of images. This work should be looked at from different perspectives. First, from a purely statistical point of view, we proved that, at least for the land cover classification problem and the chosen benchmark datasets, our models behave much better than purely propositional ones, and very good in absolute terms. Second, from a foundational standpoint spatial decision trees are located at the intersection between the symbolic learning theory and the modal logic theory, therefore opening a new field of research in which both well-known and new modal languages can be studied from a new perspective (that is, induction). Third, from a machine learning perspective, we submit that spatial symbolic learning can be a serious alternative to functional learning, typically based on neural networks, being able to offer very accurate models that can, unlike functional ones, give rise to a logical theory of the phenomena under consideration, and therefore be interpreted in a much natural way. In this respect, one should add that it is customary in machine learning to judge a new learning model uniquely from its statistical performances, which is in fact very reductive; we believe that the fact of being able to extract a theory that can be analyzed, validated, and corrected with background knowledge which can be written in the same language can compensate even the loss of some points in the accuracy of a classification exercise. Finally, our approach can be further generalized to achieve true {\em neuro-symbolic} reasoning, in which inner objects are initially recognized with a functional method (e.g., a neural network) and then their spatial relationships are exploited with a spatial decision tree. Our approach can be further developed and extended into a theory of ensemble of trees; initial experiments in this sense show that the performances of the resulting models improve in a substantial way. Ensemble of trees are usually considered to be on the verge between symbolic and functional learning: however, ensembles of trees can be interpreted, at least partially, from a logical point of view, giving rise to a logical theory of the underlying phenomenon. While computing ensembles is usually more expensive than computing single trees, the requirements can be reduced through randomization and preprocessing.
\bibliographystyle{eptcs}
\bibliography{biblio}

\end{document}